# Content Based Document Recommender using Deep Learning

Nishant Nikhil
Indian Institute of Technology Kharagpur
Kharagpur, India
Email: nishantnikhil@iitkgp.ac.in

Muktabh Mayank Srivastava
ParallelDots
Gurgaon, India
Email: muktabh@paralleldots.com

*Abstract*—With the recent advancements in information technology there has been a huge surge in amount of data available. But information retrieval technology has not been able to keep up with this pace of information generation resulting in over spending of time for retrieving relevant information. Even though systems exist for assisting users to search a database along with filtering and recommending relevant information, but recommendation system which uses content of documents for recommendation still have a long way to mature. Here we present a Deep Learning based supervised approach to recommend similar documents based on the similarity of content. We combine the C-DSSM model with Word2Vec distributed representations of words to create a novel model to classify a document pair as relevant/irrelevant by assigning a score to it. Using our model retrieval of documents can be done in O(1) time and the memory complexity is O(n), where n is number of documents.

*Keywords—Deep Learning, Recommendation system, Semantic Representation, Convolutional Neural Network*

## I. INTRODUCTION

With recent surges in information transfer, the amount of data available is increasing day by day. Due to this huge amount of available data, the need for an efficient recommendation system which can recommend data based on the similarity with the data an user interacts with is increasing. There has been a huge amount of research in this field with work on colored range queries[14] and other data structures like suffix tree[15] to efficiently search for specific patterns from the query in the documents. Word Movers' Distance[7] has also been used for retrieval purposes[16] and recently Supervised Word Movers' Distance[8] has been proposed which forces supervision on training using linear transformation.

In this study, we firstly experiment with the Convolutional Deep Structured Semantic Models (C-DSSM)[2] architecture and then tweak with its architecture to build more efficient models. The C-DSSM model evolves from DSSM (Deep Structured Semantic Model)[1] and has been used for many different tasks like information retrieval, image captioning[3], sentiment analysis[4] and a numerous tasks for natural language processing. We firstly describe the C-DSSM model then we discuss our architecture in section II after which we evaluate our model using experiments in section III by recommending research papers based on a supervised approach. Finally we conclude in section IV also mentioning about future works.

### A. Convolutional Deep Structured Semantic Models

The CDSSM model starts with feeding character n-gram based word vectors to convolutional layer which projects every word lying in a context window to a contextual vector. In the space of contextual vector, similar words-within-context get projected to nearby points. Applying it on all of the words in a sentence, the model accumulates a decent amount of local features. But the semantic meaning of a sentence generally depends only on a few key words, so it applies a max-pooling layer over it which extracts the key-global features which can effectively represent the sentence. This global vector is fixed shaped and is used as a input to a feed forward neural network which in turn extract highly effective non-linear features. Each word is represented using a tri-gram representation and each sentence with varied length vectors comprising of each word in a contextual window. Thing to notice is that even though each sentence is initially represented with varied length vectors but after max-pooling layer each sentence gets represented with fixed length vector. Therefore no padding is necessary in this model.

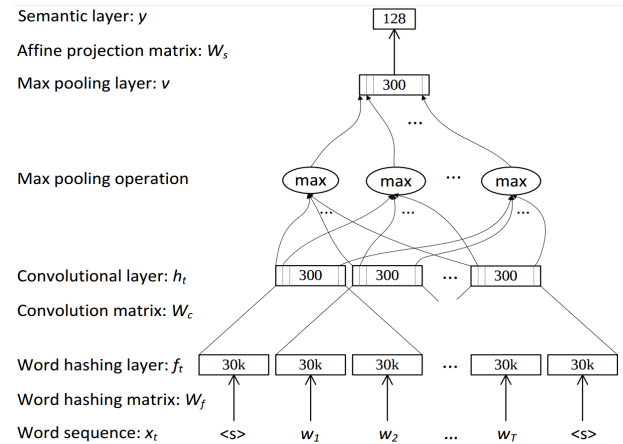

Fig. 1. CDSSM Architecture

This model is trained using the same methodology as Deep Structured Semantic Model (DSSM). In that model, two different convolutional neural networks are used to learn representation of query and documents. A relevance score is calculated between query and documents using cosine similarity as :

$$R(Q, D) = cosine(y_Q, y_D) = \frac{y_Q^T y_D}{\|y_Q\|\|y_D\|}$$

where $y_D$ and $y_Q$ are the semantic vectors of document and query respectfully. Now this is done in a supervised environment where some documents are positively related to query while some are negatively related. So the relevance score of positively related and negatively related documents are used to calculate a posterior probability of the positive document given the query by:

$$P(D^+|Q) = \frac{exp(\gamma R(Q,D^+))}{\sum_{D' \in D^-} exp(\gamma R(Q,D'))}$$

In the softmax function, $\gamma$ is a smoothing factor. $D^+$ are positively related documents and $D^-$ are negatively related documents. Then the following loss function is minimized:

$$L(\Delta) = -log \prod_{Q,D^+} P(D^+|Q)$$

where $\Delta$ denotes the model's parameters.

## II. WORKING PRINCIPLE

In this section we will describe all the components of our architecture in detail.

### A. Word vector

In the original DSSM architecture they have used character trigrams for converting a word to vector. An example of word embedding based on tri-letter concept for "cat" will be: (tri-letters are: #-c-a, c-a-t, a-t-#) This generalizes to unseen

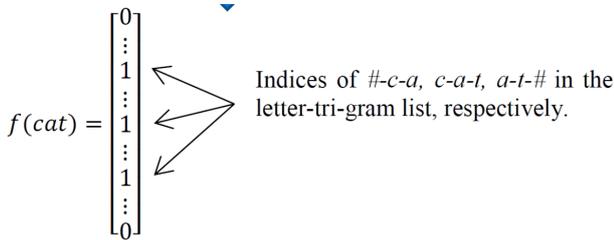

Fig. 2. Tri-Letter based representation of "cat"

word and is robust to misspelling, inflection etc. It presents a very compact way of mapping words. In their experiment they found a vocabulary of 500,000 being mapped to 30,000 dimensional vector. But there is a possibility of collision. For different experiments they have shown:

Table1 : Number of Collisions

| Vocabulary Size | Unique hashed vectors | Number of collisions |
|---|---|---|
| 40K | 10306 | 2 |
| 500K | 30621 | 22 |

These collisions are too small in number that we can neglect them.

In our model when we improve on these word embedding, we will use word embeddings obtained by applying WORD2VEC on abstracts from PubMed. This follows from [6] which advocates that unsupervised training of word vectors can hugely improve a NLP model.

### B. Sentence Embeddings

We use the same approach used by original DSSM paper. We concatenate word embeddings of previous k words and next k words for every word to build a context window of 2k+1 words. For example sentence embedding for "Ram loves playing cricket" will be {vec(Ram) + vec(loves) + vec(playing), vec(loves) + vec(playing) + vec(cricket)} in a context window of three words, where + represents concatenation operation.

### C. Convolutional feature Maps

A convolutional layer extracts pattern from input data. Here it extracts discriminative word sequences present in the input sentence. Mathematically it can be represented as:

$$c_i = (S * F)_i = \sum_{k,j}(S_{[:,i-m+1:i]} \otimes F)_{kj}$$

Where the input is $S \in R^{dxs}$ and F is a filter $\in R^{dxm}$ of width m. It results in a vector $c \in R^{s+m-1}$. Here $\otimes$ is the elements wise multiplication and $S_{[:,i-m+1:i]}$ denotes a slice of size m along the columns.

### D. Activation Units

To learn non-linear feature from the input data, the convolutional layer is followed by a non-linear activation function which is applied element wise. We have used hyperbolic tangent function in our model.

$$\tanh x = \frac{\sinh x}{\cosh x} = \frac{e^{2x}-1}{e^{2x}+1}$$

### E. Max-Pooling Layer

The Max-Pooling layer extracts the maximum activated feature for all the convolutional filters. In this way we are extracting the key-features from the data. Though this can be extended by using k-max pooling[5], which extracts k most activated features. This can be used to make model deeper but we only use Max-Pooling as we needed a model which can train and run fast and [6] advocates that single-layer networks perform equally well.

### F. Tweaked Model

We tweaked with the architecture and came up with two modifications, firstly we replaced the two different convolutional neural network model for query and documents to one common model for both the query and documents. In the original C-DSSM paper the query and documents came from different sources but in our case we have both the query and documents from same data so using one single convolutional neural network not only reduced the training parameters by half, but also reduced the training time by more than one fourth of original training time. And also increased the accuracy in our case.

Then we use the 200-Dimensional word embeddings obtained by applying WORD2VEC[9] to approximately 11 million abstracts from PubMed. This is provided by BioASQ. We deduce that in the original C-DSSM paper tri gram representation were necessary to generalise unseen words as the experiment is done in the web environment where you can't have a bounded vocabulary, but it is not the case here.

As we had to use the model for production purposes we wanted the model to be accurate and fast at the same time. Though the paper on Word Movers' Distance[7] and its supervised extension, Supervised Word Movers' Distance[8] show promising results but the speed of model is questionable here. Using this tweaked model of C-DSSM doesn't only give us accuracy but speed also as each of the document can be hashed using the model and at run-time we only need to find its k-nearest neighbour which can be done in O(1) following the same approach as [13].

For a better intuition we can think of original C-DSSM learning two different convolutional networks because in the task at hand the query space generally has at most one sentence (web-search query) while the fetched documents might have more than one sentence. So there should be different patterns we are looking in both the spaces. There are lesser combinations of different pattern in query space, so the query space is highly polarised through specific keywords. Due to this we need to learn different weights for both query space and search space. But when we have the same query and search space, we are looking for same patterns in query and search space. This can be further be clarified by taking the example of a convolutional neural networks trained only on big images of object, it usually fares poorly on detecting small images of same object. As is the case of detecting logo in DeepLogo[10] which although works great on images having high resolution logos but fares poorly on images containing low resolution logos.

About using pre-trained word vectors, whilst in the original C-DSSM paper they had to generalize about unseen words, spelling mistakes and other alterations but in our case the vocabulary is well-structured and it is rare to find an unseen word so using pre-trained vectors which has been shown in [11][12] to improve accuracy resulted the same way in our case too. The pre-trained vectors which are trained in an unsupervised manner generally captures some semantic and syntactic relation between the words.

III. EXPERIMENTS

We used the data from BioASQ 2016 Task 4b in which questions along with their relevant documents are provided. We use this as a basis to categorize positively related documents. All the documents related to a question are in turn categorized as positively related to each other while all other documents are considered negatively related to each document related to the question. Then we build a separated train and validation set. All the metrics in the paper are calculated for validation set.

As a baseline approach we used Latent Dirichlet Allocation to get vector representation of documents. We got the following results for it:

Table2 :  LDA based results

| K | recall@k | precision@k | f1@k |
|---|---|---|---|
| 1 | 0.045 | 0.045 | 0.045 |
| 5 | 0.044 | 0.043 | 0.044 |
| 10 | 0.045 | 0.041 | 0.043 |
| 20 | 0.050 | 0.038 | 0.043 |
| 30 | 0.062 | 0.037 | 0.047 |
| 40 | 0.076 | 0.037 | 0.050 |
| 50 | 0.090 | 0.036 | 0.052 |

For an un-altered C-DSSM model we got the following recall and precision:

Table3 :  Un-altered C-DSSM

| K | recall@k | precision@k | f1@k |
|---|---|---|---|
| 1 | 0.190 | 0.190 | 0.190 |
| 5 | 0.179 | 0.176 | 0.177 |
| 10 | 0.178 | 0.163 | 0.170 |
| 20 | 0.195 | 0.196 | 0.196 |
| 30 | 0.226 | 0.135 | 0.169 |
| 40 | 0.259 | 0.125 | 0.169 |
| 50 | 0.291 | 0.119 | 0.169 |

After combining the two convolutional models into one, we got the following results:

Table4 :  Model with common convolutional layer

| K | recall@k | precision@k | f1@k |
|---|---|---|---|
| 1 | 0.283 | 0.283 | 0.283 |
| 5 | 0.267 | 0.262 | 0.264 |
| 10 | 0.265 | 0.244 | 0.254 |
| 20 | 0.291 | 0.218 | 0.249 |
| 30 | 0.333 | 0.199 | 0.249 |
| 40 | 0.377 | 0.183 | 0.247 |
| 50 | 0.418 | 0.170 | 0.242 |

After incorporating the pre-trained word vectors we got even better results:

Table5 :  Model with common convolutional layer and pretrained word embeddings

| K | recall@k | precision@k | f1@k |
|---|---|---|---|
| 1 | 0.411 | 0.410 | 0.410 |
| 5 | 0.378 | 0.370 | 0.374 |
| 10 | 0.370 | 0.340 | 0.354 |
| 20 | 0.399 | 0.299 | 0.342 |
| 30 | 0.445 | 0.266 | 0.333 |
| 40 | 0.494 | 0.240 | 0.323 |
| 50 | 0.538 | 0.220 | 0.311 |

To compare thewse models we have:

Table6 :  Comparisons

| K | LDA | Un-altered C-DSSM | Common convolutional layer | Common convolutional layer with pretrained word embeddings |
|---|---|---|---|---|
| f1@1 | 0.045 | 0.190 | 0.283 | 0.410 |
| f1@5 | 0.044 | 0.177 | 0.264 | 0.374 |
| f1@10 | 0.043 | 0.170 | 0.254 | 0.354 |

Here recall is the number of documents which are retrieved and are relevant divided by number of documents which are relevant. And precision is the number of documents which are retrieved and are relevant divided by number of documents which are retrieved. To sum up:

$$recall = \frac{|\{relevant\ documents\} \cap \{retrieved\ documents\}|}{|\{relevant\ documents\}|}$$
$$precision = \frac{|\{relevant\ documents\} \cap \{retrieved\ documents\}|}{|\{retrieved\ documents\}|}$$

Now, recall@k is defined as proportion of relevant documents that are in top k. And precision@k is defined as proportion of top k documents that are relevant. But in the data, for each document we had different number of positively labeled documents. So for calculating recall@k when k is less than the number of relevant documents, we take k as the number of relevant documents. This is done to avoid getting a pessimistic view of the model.

## IV. CONCLUSION

From our experiments we see that using single neural network to learn the embeddings of documents did not only reduce the complexity but also increased the accuracy. Intuitively we see that as the space for query and fetched documents are same, we need the convolutional layer to learn features which is common to both. Then our model adds to the previous belief that unsupervised training of word vectors can hugely improve a NLP model.

## REFERENCES

[1] Huang, Po-Sen, Xiaodong He, Jianfeng Gao, Li Deng, Alex Acero, and Larry Heck. "Learning deep structured semantic models for web search using clickthrough data." In Proceedings of the 22nd ACM international conference on Conference on information & knowledge management, pp. 2333-2338. ACM, 2013.

[2] Shen, Yelong, Xiaodong He, Jianfeng Gao, Li Deng, and Grgoire Mesnil. "A latent semantic model with convolutional-pooling structure for information retrieval." In Proceedings of the 23rd ACM International Conference on Conference on Information and Knowledge Management, pp. 101-110. ACM, 2014.

[3] Fang, Hao, Saurabh Gupta, Forrest Iandola, Rupesh K. Srivastava, Li Deng, Piotr Dollr, Jianfeng Gao et al. "From captions to visual concepts and back." In Proceedings of the IEEE Conference on Computer Vision and Pattern Recognition, pp. 1473-1482. 2015.

[4] Severyn, Aliaksei, and Alessandro Moschitti. "Twitter sentiment analysis with deep convolutional neural networks." In Proceedings of the 38th International ACM SIGIR Conference on Research and Development in Information Retrieval, pp. 959-962. ACM, 2015.

[5] Kalchbrenner, Nal, Edward Grefenstette, and Phil Blunsom. "A convolutional neural network for modelling sentences." arXiv preprint arXiv:1404.2188 (2014).

[6] Kim, Yoon. "Convolutional neural networks for sentence classification." arXiv preprint arXiv:1408.5882 (2014).

[7] Kusner, Matt J., Yu Sun, Nicholas I. Kolkin, and Kilian Q. Weinberger. "From Word Embeddings To Document Distances." In ICML, vol. 15, pp. 957-966. 2015.

[8] Huang, Gao, Chuan Guo, Matt J. Kusner, Yu Sun, Fei Sha, and Kilian Q. Weinberger. "Supervised Word Mover's Distance." In Advances in Neural Information Processing Systems, pp. 4862-4870. 2016.

[9] Mikolov, Tomas, Kai Chen, Greg Corrado, and Jeffrey Dean. "Efficient estimation of word representations in vector space." arXiv preprint arXiv:1301.3781 (2013).

[10] Iandola, Forrest N., Anting Shen, Peter Gao, and Kurt Keutzer. "DeepLogo: Hitting logo recognition with the deep neural network hammer." arXiv preprint arXiv:1510.02131 (2015).

[11] Chen, Danqi, and Christopher D. Manning. "A Fast and Accurate Dependency Parser using Neural Networks." In EMNLP, pp. 740-750. 2014.

[12] Socher, Richard, Danqi Chen, Christopher D. Manning, and Andrew Ng. "Reasoning with neural tensor networks for knowledge base completion." In Advances in neural information processing systems, pp. 926-934. 2013.

[13] Krizhevsky, Alex, and Geoffrey E. Hinton. "Using very deep autoencoders for content-based image retrieval." In ESANN. 2011.

[14] Gagie, Travis, Juha Krkkinen, Gonzalo Navarro, and Simon J. Puglisi. "Colored range queries and document retrieval." Theoretical Computer Science 483 (2013): 36-50.

[15] Muthukrishnan, S. "Efficient algorithms for document retrieval problems." In Proceedings of the thirteenth annual ACM-SIAM symposium on Discrete algorithms, pp. 657-666. Society for Industrial and Applied Mathematics, 2002.

[16] Brokos, Georgios-Ioannis, Prodromos Malakasiotis, and Ion Androutsopoulos. "Using centroids of word embeddings and word mover's distance for biomedical document retrieval in question answering." arXiv preprint arXiv:1608.03905 (2016).